\documentclass[conference]{IEEEtran}
\IEEEoverridecommandlockouts
\usepackage{cite}
\usepackage{amsmath,amssymb,amsfonts}
\usepackage{algorithmic}
\usepackage{graphicx}
\usepackage{textcomp}
\usepackage{xcolor}
\usepackage{subcaption}
\usepackage{booktabs}
\usepackage{hyperref}
\usepackage{float,orcidlink}

\makeatletter
\newcommand{\linebreakand}{%
  \end{@IEEEauthorhalign}
  \hfill\mbox{}\par
  \mbox{}\hfill\begin{@IEEEauthorhalign}
}
\makeatother

\def\BibTeX{{\rm B\kern-.05em{\sc i\kern-.025em b}\kern-.08em
    T\kern-.1667em\lower.7ex\hbox{E}\kern-.125emX}}
\begin{document}

\title{Impact of spatial transformations on landscape features of CEC2022 basic benchmark problems
}

\author{\IEEEauthorblockN{1\textsuperscript{st} Haoran Yin}
\IEEEauthorblockA{\textit{LIACS, Leiden University}\\
Leiden, The Netherlands \\
\url{h.yin@liacs.leidenuniv.nl}~\orcidlink{0009-0005-7419-7488}}
\and
\IEEEauthorblockN{2\textsuperscript{nd} Diederick Vermetten}
\IEEEauthorblockA{\textit{LIACS, Leiden University}\\
Leiden, The Netherlands \\
\url{d.l.vermetten@liacs.leidenuniv.nl}~\orcidlink{0000-0003-3040-7162}}
\and
\IEEEauthorblockN{3\textsuperscript{rd} Furong Ye}
\IEEEauthorblockA{\textit{
ISCAS, Chinese Academy of Science}\\
Beijing, China\\
\url{f.ye@ios.ac.cn}~\orcidlink{0000-0002-8707-4189}}
\linebreakand
\IEEEauthorblockN{4\textsuperscript{th} Thomas H.W. B{\"a}ck}
\IEEEauthorblockA{\textit{LIACS, Leiden University}\\
Leiden, The Netherlands \\
\url{t.h.w.baeck@liacs.leidenuniv.nl}~\orcidlink{0000-0001-6768-1478}}
\and
\IEEEauthorblockN{5\textsuperscript{th} Anna V. Kononova}
\IEEEauthorblockA{\textit{LIACS, Leiden University}\\
Leiden, The Netherlands \\
\url{a.kononova@liacs.leidenuniv.nl}~\orcidlink{0000-0002-4138-7024}}
}


\maketitle

\begin{abstract}

When benchmarking optimization heuristics, we need to take care to avoid an algorithm exploiting biases in the construction of the used problems. One way in which this might be done is by providing different versions of each problem but with transformations applied to ensure the algorithms are equipped with mechanisms for successfully tackling a range of problems. In this paper, we investigate several of these problem transformations and show how they influence the low-level landscape features of a set of 5 problems from the CEC2022 benchmark suite. Our results highlight that even relatively small transformations can significantly alter the measured landscape features. This poses a wider question of what properties we want to preserve when creating problem transformations, and how to fairly measure them.  

\end{abstract}


\section{Introduction}
\label{intro:background}

In recent decades, numerous optimization algorithms have been developed \cite{back2023evolutionary, zhang2015comprehensive}.
According to the no-free-lunch-theorem \cite{585893}, none of these algorithms can be dominant on all optimization problems, which means that some algorithms will perform better than others on specific problems. It is not easy to determine the conditions under which optimization algorithms perform well, and rigorously benchmarking the algorithms is a common way to address this \cite{bartzbeielstein2020benchmarking}. Benchmarking should encompass a broad spectrum of representative functions, with an emphasis on generating multiple instances of each function to reduce bias, improve 
robustness, better simulate 
real-world conditions, and encourage  
the development of more versatile and adaptive algorithms \cite{bartzbeielstein2020benchmarking, bartz2010sequential, whitley1996evaluating}. The mechanism for generating instances should maintain the fundamental landscape structure and attributes of the original function while introducing variations, such as shifts in the optima locations and changes in function value amplitudes. 
This approach prevents the optimization algorithm design from becoming too specific for a specific function landscape, or from benefiting from a strong structural bias towards specific regions of the search space~\cite{vermetten2022bias, kudela2022critical}.

Different instances of the same underlying problem can be created in a variety of ways. For example, in pseudo-boolean optimization, variables might be shifted and then fed through an XOR with a random bitstring~\cite{LehreW12}; such transformations have been applied for the pseudo-boolean optimization suite of the IOHprofiler benchmark environment~\cite{de2021iohexperimenter}. Applying these transformations to the well-known OneMax problem efficiently removes the specific bias towards the value of 1 while maintaining the problem structure. 
In real-valued optimization, problem instances are generally created by applying a set of transformations to a base problem. This is the approach taken by the BBOB suite, which is one of the most well-established sets of benchmark problems in continuous, noiseless optimization~\cite{bbobfunctions, hansen2021coco}. By generating seeded scaling, rotation and translation methods, the global landscape properties of the base functions are preserved, which then allows for the testing of several algorithm invariances~\cite{hansen2011impacts}.

While the transformation methods used in instance generation are generally designed to preserve high-level problem properties, their exact impact on the low-level landscape is obvious. From the perspective of Exploratory Landscape Analysis (ELA), the different box-constrained BBOB instances are statistically different in a variety of ways, and corresponding algorithm performance can vary as a result. 

To better understand the relation between problem transformations and landscape features, we use another popular set of continuous black-box optimization problems, known as the CEC2022 problem suite. Unlike the BBOB suite, the CEC2022 suit does not natively support instance generation. As such, it provides an ideal testbed for the study of various transformation methods, which might help in determining useful guidelines for future instance generation within this problem suite. 

The remainder of this paper is structured as follows: Section~\ref{sec:related work} provides an overview of relevant previous research, with a focus on landscape features. This section also describes the CEC2022 problem suite. In Section \ref{chap:ExperimentalSetup}, we introduce our experimental setup, which includes the specific ELA 
 features used and the full set of problem transformations we consider. Results are then discussed in Section \ref{sec:results}, after which Section \ref{sec:conclusion} discusses the key conclusions and highlights possible future work.

\begin{figure*}[]
    \centering
    \begin{subfigure}[b]{0.3\linewidth}
        \includegraphics[width=\linewidth]{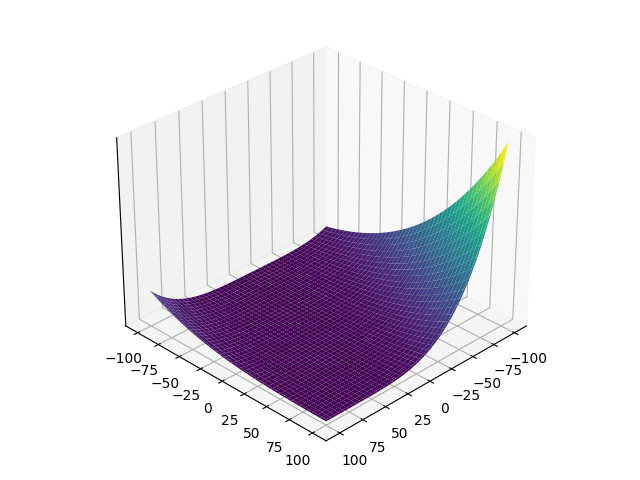}
        \caption{problem 1, Zakharov Function \cite{floudas2013handbook}}
        \label{fig:CEC2022_landscape_subfig1}
    \end{subfigure}
    \begin{subfigure}[b]{0.3\linewidth}
        \includegraphics[width=\linewidth]{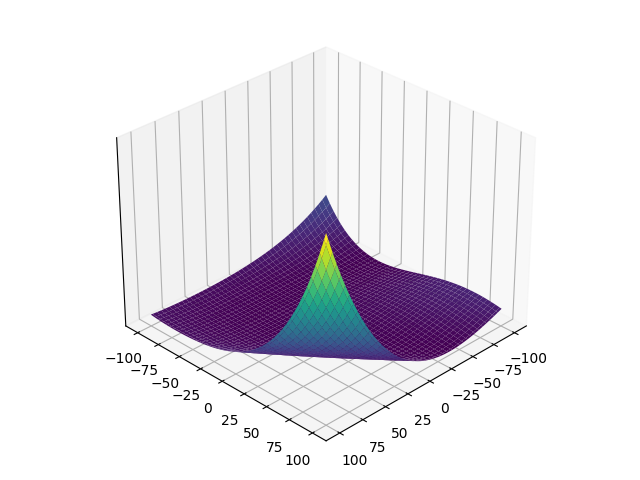}
        \caption{problem 2, Rosenbrock’s Function \cite{rosenbrock1960automatic}}
        \label{fig:CEC2022_landscape_subfig2}
    \end{subfigure}
    \begin{subfigure}[b]{0.3\linewidth}
        \includegraphics[width=\linewidth]{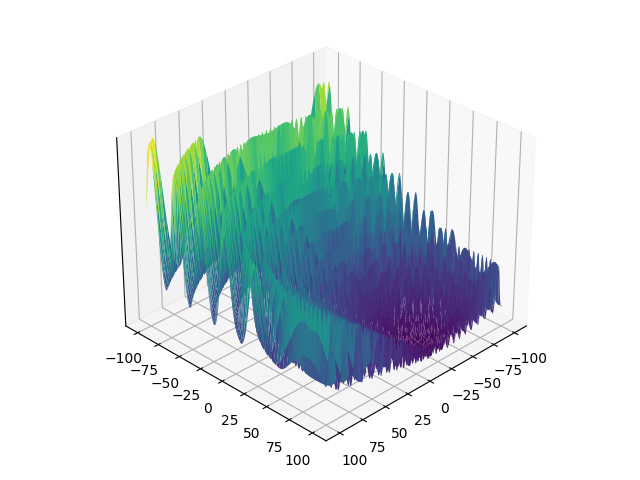}
        \caption{problem 3, Schaffer’s F7 Function \cite{schaffer2014multiple}}
        \label{fig:CEC2022_landscape_subfig3}
    \end{subfigure}
    \begin{subfigure}[b]{0.3\linewidth}
        \includegraphics[width=\linewidth]{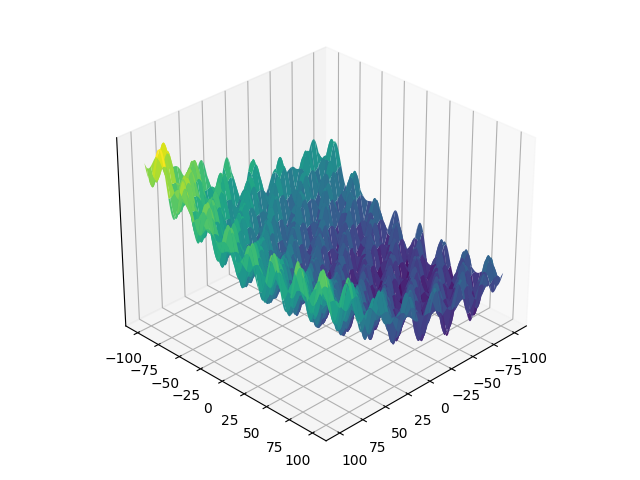}
        \caption{problem 4, Rastrigin’s Function \cite{beyer2002evolution}}
        \label{fig:CEC2022_landscape_subfig4}
    \end{subfigure}
    \begin{subfigure}[b]{0.3\linewidth}
        \includegraphics[width=\linewidth]{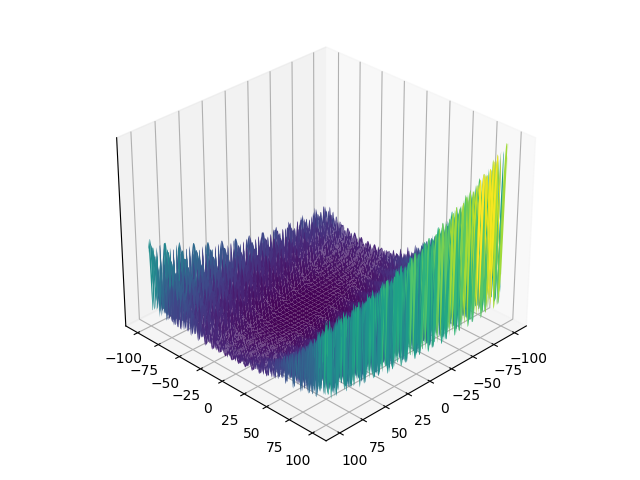}
        \caption{problem 5, Levy Function \cite{floudas2013handbook}}
        \label{fig:CEC2022_landscape_subfig5}
    \end{subfigure}
    \caption{Landscapes of CEC2022 basic problems in $[-100, 100]^2$. }
    \label{fig:CEC2022_landscape}
\end{figure*}

\section{Related Work} \label{sec:related work}
In this section, we explore existing work, outline several key studies within the field and discuss their relevance to this work.

The methodology of Exploratory Landscape Analysis (ELA) was introduced for characterizing the properties of the objective function landscape \cite{mersmann2011exploratory} to potentially facilitate the recommendation of well-performing algorithms for unseen problems. One possible way to achieve this is to understand how problem properties influence algorithm performance and group test problems into classes with similar performance of the optimization algorithms. ELA was proposed to solve this based on some numerical features (relatively) cheaply computed from limited samples from the function landscape. With time, ELA has evolved into an umbrella term for analytical, approximated and non-predictive methods covering a wide range of characteristics of function landscapes~\cite{munoz2015alg}. While it has been previously shown that no single exact or approximate easily computable proxy of function difficulty is possible for black-box optimization~\cite{he2007note}, typical modern usage of ELA employs multiple features to characterize the landscape in aspects such as convexity, function values distribution, curvature, meta-model and local search features, dispersion, information content and principle component features, to name a few~\cite{mersmann2011exploratory,KerschkeT2019flacco}. 

The state-of-the-art application of ELA analysis requires careful consideration of the number of questions: 
\begin{itemize}
    \item Since typically, explicit problem representation is not available, feature values need to be estimated based on a small number of sample points. Informed answers are needed to questions like what sampling strategy should be used (structured vs unstructured~\cite{munoz2015alg}, random vs Latin hypercube vs low discrepancy-sequence~\cite{renau2020exploratory}) and how many points need to be considered for a general problem dimensionality to obtain robust estimates cheaply~\cite{renau2020exploratory}.
    \item It has been shown that the set of landscape features provided by popular ELA libraries is highly redundant~\cite{vskvorc2020understanding}, especially for higher problem dimensionality~\cite{renau2021towards}. This necessitates efficient feature selection methods. 
    \item The additional benefits from more complex features need to be balanced against high computation time which furthermore does not scale favourably with problem dimensionality~\cite{munoz2015alg}.
    \item Computed values of some features require careful preprocessing since they might not be invariant to, e.g., scaling of the objective function~\cite{skvorc2021effect}. 
\end{itemize}


Furthermore, in a recent study, Long et al.~\cite{long2023bbob} used ELA to investigate the landscape characteristics of BBOB problem instances and the instance generation process by analyzing 500 instances of each BBOB problem. 
The experiments reveal a large diversity in the distributions of ELA features, even for instances of the same BBOB function. 
Furthermore, the authors tested the performance of eight algorithms on these 500 instances and investigated statistically significant differences between the performances. 
The article asserts that although the transformations applied to the BBOB instances preserve the high-level properties of the functions, in practice these differences should not be ignored, especially when the problem is treated as bounded constraints rather than unconstrained.
Skvorc et al.~investigated the resilience of ELA features against basic function transformations such as shifting and scaling \cite{vskvorc2022comprehensive}. 
They find that some of the ELA features remain robust under various conditions, supporting their reliability in automating algorithm selection processes, despite some sensitivity in specific instances. 
This work underscores the importance of ELA in characterizing optimization problems and guiding the choice of the algorithm.



\begin{figure*}[]
    \centering
    \includegraphics[width=\linewidth]{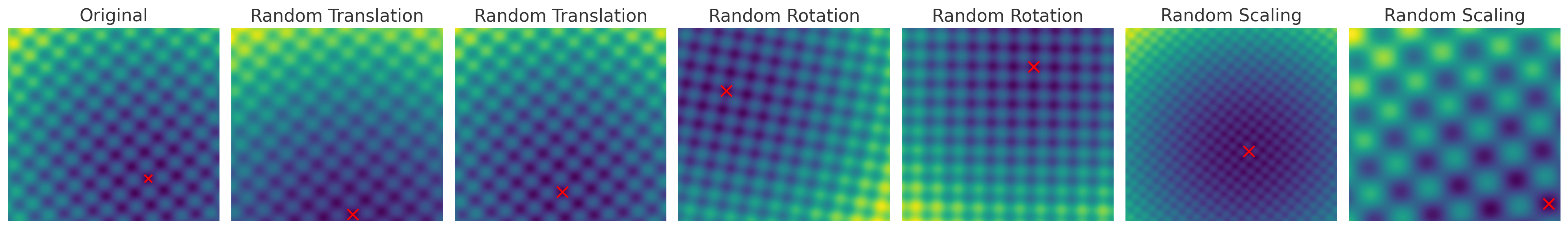}
    \caption{Examples of instance generation via spatial transformations applied to one of the original CEC2022 benchmark problems, shown here in two dimensions, with optima locations marked by crosses. }
    \label{fig:impact_transformation}
\end{figure*}

In light of the above research, our work aims to analyze transformations which could potentially be used in an instance generation system for the basic CEC2022 problems \cite{ahrari2022problem}. 
Fig.~\ref{fig:CEC2022_landscape} shows the landscape of CEC2022 basic problems in a 2D search space, and
Fig.~\ref{fig:impact_transformation} shows three types of transformation that apply to the search space.
Although there have been many studies on the CEC2022 problems, its generation of instances still lacks in-depth exploration \cite{vskvorc2022comprehensive}. 
This is because officially only one instance is provided for each problem in the competition \cite{ahrari2022problem}, and researchers are hardly exploring the generation of other instances. 
The above fact motivates our research on the instance generation system and on investigating the impact of spatial transformations on landscape features for the CEC2022 basic benchmark problems.



\section{Experimental Setup}
\label{chap:ExperimentalSetup}

\subsection{ELA features used in the experiments} 
\label{ExperimentalSetup:landscape features}
\textit{flacco} is an \textbf{R}-package that provides an implementation of the ELA feature calculation \cite{KerschkeT2019flacco}. 
It provides a number of feature sets that can compute values based on relatively small samples of the search space, thus describing the broader and more specific characteristics of the problem.
From \textit{flacco}, we choose a set of 55 features widely used by researchers \cite{long2023bbob}. 

\subsection{Experiment data}
\label{ExperimentalSetup:SpatialTransformations}
Let $\boldsymbol{x}$ be the solution in the search space and $y = f(\boldsymbol{x})$ 
be its objective function value. 
In the following, we provide the definitions of
considered spatial transformations and the method for collecting data from the spatial transformation experiment. The settings provided below are tailored for the CEC functions which are defined in $[-100,100]^{10}$.

\subsubsection{Transformations on the search space}
\begin{itemize}
    \item \textbf{Translation}: For every $i$-th component of $\boldsymbol{x}$, a translation offset is independently sampled from $\mathcal{U}(-d_{\boldsymbol{x}},d_{\boldsymbol{x}})$ and added to $\boldsymbol{x}_i$, to generate $\boldsymbol{x}'$. To examine the influence of a translation on the search space, multiple experiments are carried out with $d_{\boldsymbol{x}} \in D_{\boldsymbol{x}}=\left\{ 5, 10, 15, ..., 100 \right\}$, where for each translation limit, 10 random translation vectors are generated to obtain the averaged results. 
    \item \textbf{Scaling}: For the solution $\boldsymbol{x}$ from the search space, $\boldsymbol{x}'=k_{\boldsymbol{x}}\boldsymbol{x}$, where $k_{\boldsymbol{x}}$ is a scaling factor .
    To fully study this transformation, a number of scaling factors is considered from $K_{\boldsymbol{x}} = \left\{ 2^{-6}, 2^{-5}, ..., 2^{5}, 2^{6} \right\}$, to record their influence,
    which allows us to explore how different levels of scaling in the search space affect the ELA features.
    Currently, these factors are enough for us to study the influence of scaling and give a concrete discussion in Section \ref{sec:results}. Furthermore, factors that are smaller than $2^{-6}$ make the search space for CEC2022 basic problems too small.
    
    \item \textbf{Rotation}: The rotation matrix is generated at random and applied to the input variable $\boldsymbol{x}$; this setup is repeated 30 times. Such an approach is preferred over investigating rotations at various angles since the latter is difficult to interpret in high-dimensional spaces. 
\end{itemize}

\subsubsection{Transformations on the objective value}
\begin{itemize}
    \item \textbf{Objective translation}: For the objective value $y$, a translation offset $d_y$ is added to $y$, to generate $y'$. To investigate the impact of translation on the objective value, various experiments are carried out with 10 translation values $ d_y \in D_y = \left\{ 100, 200, ..., 1000 \right\}$. 
    \item \textbf{Objective scaling}: For the objective value $y$, a scaling factor $k_{y}$ is multiplied by $y$, to generate $y'$.
    To study its influence, a set of scaling factors $K_{y} = \left\{ 2^{-6}, 2^{-5}, ..., 2^{5}, 2^{6} \right\}$ is applied based on the experiments with scaling via $k_{x}$.
\end{itemize}

\subsubsection{Data collection}

While the CEC2022 competition, which is the source of the functions we use, encompassed several different dimensionalities ($d\in\{2, 10, 20\}$), we focus only on the 10-dimensional version of the suite, which has a domain of  $[-100, 100]^{10}$. 

In total, $267$ transformations (`instances') are considered here per function: 1 original, 200 
with translated search space, 30 with rotated search space, 13 with scaled search space, 10 with translated objective values and 13 with scaled objective values. 

On each instance, ELA features are computed using \textit{flacco} based on $m=100\cdot d$ points produced by Latin hypercube sampling \cite{eglajs1977new}. This value was chosen to maintain a balance between computation time and feature stability~\cite{renau2019expressiveness}. However, since this sampling-based process is by definition stochastic 
, we repeat the sampling $100$ times for each generated function instance. These ELA features are then normalized. Given that we make use of a total of 55 ELA features
, we end up with a set of $267\cdot55\cdot100=1\,468\,500$ feature values, per base function. The detailed data from the experiment are open for access \cite{anonymous2024impact}.

\subsection{Methodology}
\label{sec:methods}
\subsubsection{Dimensionality reduction}
\label{sec:umap}
Uniform Manifold Approximation and Projection (UMAP) is an algorithm for dimensionality reduction and visualization of high-dimensional data \cite{mcinnes2018umap}. 
It helps us understand and analyze complex datasets by mapping the data to the manifold in a space with lower dimensionality and preserving the local structure from high-dimensional space. 
We apply this algorithm for mapping results from the 55-dimensional ELA feature space to a 2D space, 
to better understand how spatial transformations influence the ELA features of the CEC2022 problems.

\subsubsection{Statistical test}
\label{sec:ks-test}
The Kolmogorov-Smirnov test (KS-test), a non-parametric statistical test, determines whether two distributions are statistically the same \cite{kolmogorov1933sulla}. 
Among the statistical results of the KS-test, we consider $p\text{-value}$ to be the most important. 
As two sets of samples are denoted by symbols $P$ and $Q$, if $p\text{-value} > \alpha$, the observed distinction between $P$ and $Q$ does not have a statistically significant impact (here, $\alpha=0.05$).
Otherwise, the hypothesis that $P$ and $Q$ conform to the same distribution is rejected by the KS-test.
The KS-test helps assess whether the spatial transformation has an impact on the ELA feature and how this impact changes as the transformation level changes.

\subsubsection{Difference measure}
\label{sec:emd}
Earth Mover's Distance (EMD), a concept used interchangeably with the Wasserstein metric, quantifies the difference between the two distributions \cite{kantorovich1960mathematical}. 
It represents the minimum cost required to move the mass from one distribution to another. 
We use this measure to have a clearer understanding of how spatial transformation affects different benchmark problems and to contrast it with the KS-test, helping to draw further conclusions.








\section{Results: Impact of transformations}
\label{sec:results}

After obtaining the experimental data, we applied UMAP (see Section \ref{sec:umap}) to the data of the ELA features, represented as $55$-dimensional vectors, to a 2-dimensional projection. The projection mapping is created on the original functions and then applied to all constructed instances of all functions. 

\begin{figure*}[!tb]
    \centering
    \begin{subfigure}[b]{0.45\linewidth}
        \includegraphics[width=\linewidth]{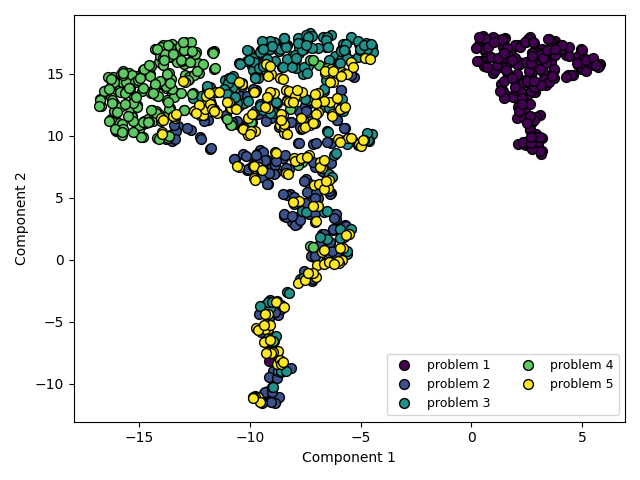}
        \caption{Original 5 basic problems}
        \label{fig:search_space_2d}
    \end{subfigure}
    \begin{subfigure}[b]{0.45\linewidth}
        \includegraphics[width=\linewidth]{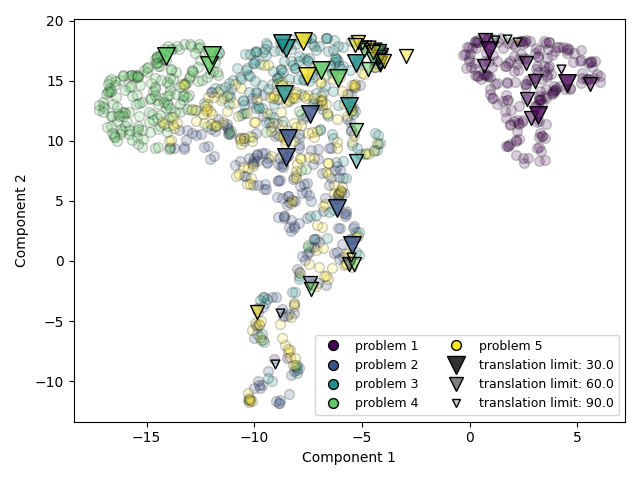}
        \caption{Before and after translation on $\boldsymbol{x}$}
        \label{fig:search_space_2d_subtract}
    \end{subfigure}
    \begin{subfigure}[b]{0.45\linewidth}
        \includegraphics[width=\linewidth]{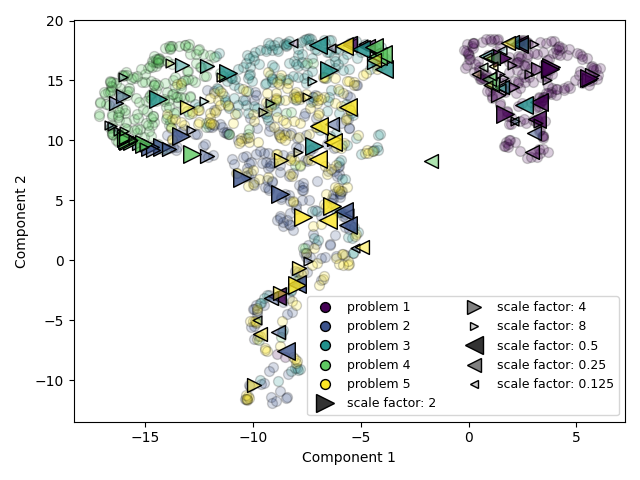}
        \caption{Before and after scaling on $\boldsymbol{x}$}
        \label{fig:search_space_2d_scale}
    \end{subfigure}
    \begin{subfigure}[b]{0.45\linewidth}
        \includegraphics[width=\linewidth]{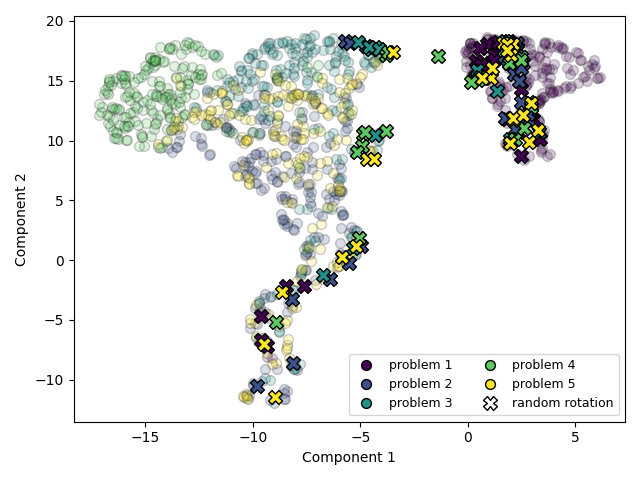}
        \caption{Before and after rotation on $\boldsymbol{x}$}
        \label{fig:search_space_2d_rotate}
    \end{subfigure}
    \begin{subfigure}[b]{0.45\linewidth}
        \includegraphics[width=\linewidth]{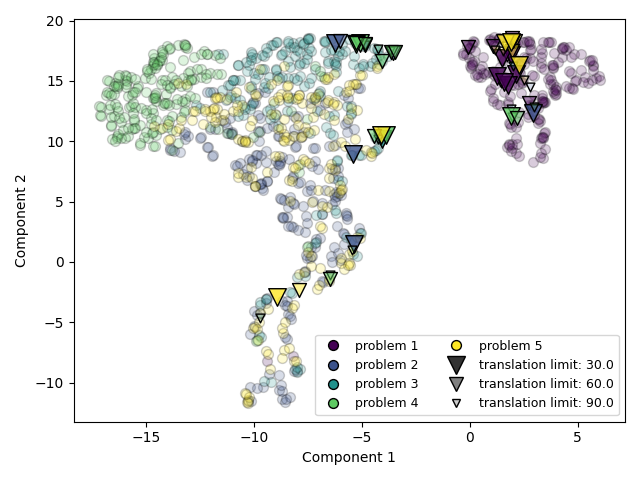}
        \caption{Before and after translation on $y$}\label{fig:objective_value_2d_subtract}
    \end{subfigure}
    \begin{subfigure}[b]{0.45\linewidth}
        \includegraphics[width=\linewidth]{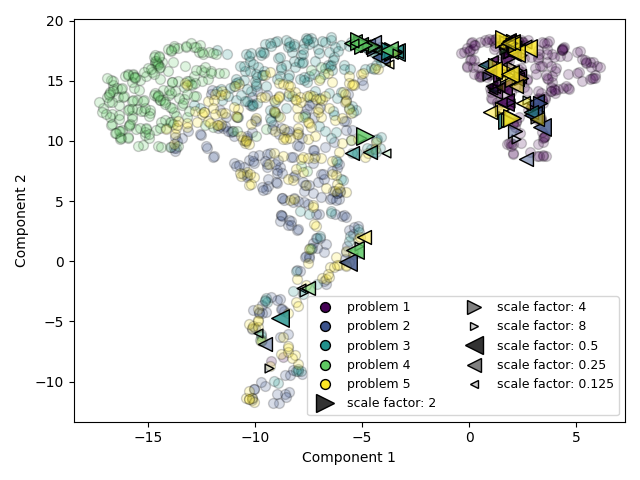}
        \caption{Before and after scaling on $y$}\label{fig:objective_value_2d_scale}
    \end{subfigure}
    \caption{
    UMAP-based projections of ELA features for various spatial transformations of five basic problems. The projection model is computed based on original problems and is applied to all samples shown in these figures. 100 samples of each original problem are shown with round (faded) coloured symbols, while samples of the transformed problems correspond to other coloured shapes, with different transformation magnitudes denoted by different symbol sizes. Each point represents a set of ELA features calculated after a Latin hypercube sampling in the search space. Only randomly subselected 4\% of the transformed problem samples are presented in the graph to avoid overwhelming information with large data volumes.}
    \label{fig:umap_search_space}
\end{figure*}

Fig. \ref{fig:search_space_2d} shows the resulting scatter plot of the ELA features of the five benchmark problems. Each point represents a projection of the full ELA feature vector calculated on a Latin hypercube sampling, while different colours represent different base problems. It appears that problem 1 before the generation of instances has a high degree of separation from other problems, which shows that in the ELA feature space, the features of problem 1 are significantly different from those of other problems. Problems 2 to 5 have a less clear separation in the projected space, although differences between problems 3 and 4 are still easy to identify. 

To illustrate the impact of the various transformation methods discussed in Section~\ref{chap:ExperimentalSetup}, Figures \ref{fig:search_space_2d_subtract} to \ref{fig:objective_value_2d_scale} show the location of the transformed problems under the same mapping. This highlights the extent to which these transformations change the overall ELA feature vectors. Throughout the remainder of this section, we will zoom in on each transformation method to identify the relation between its parameterization and the change in ELA representation of the resulting function.


\begin{figure*}[!t]
    \centering
    \begin{subfigure}[b]{0.48\linewidth}
        \includegraphics[width=\linewidth]{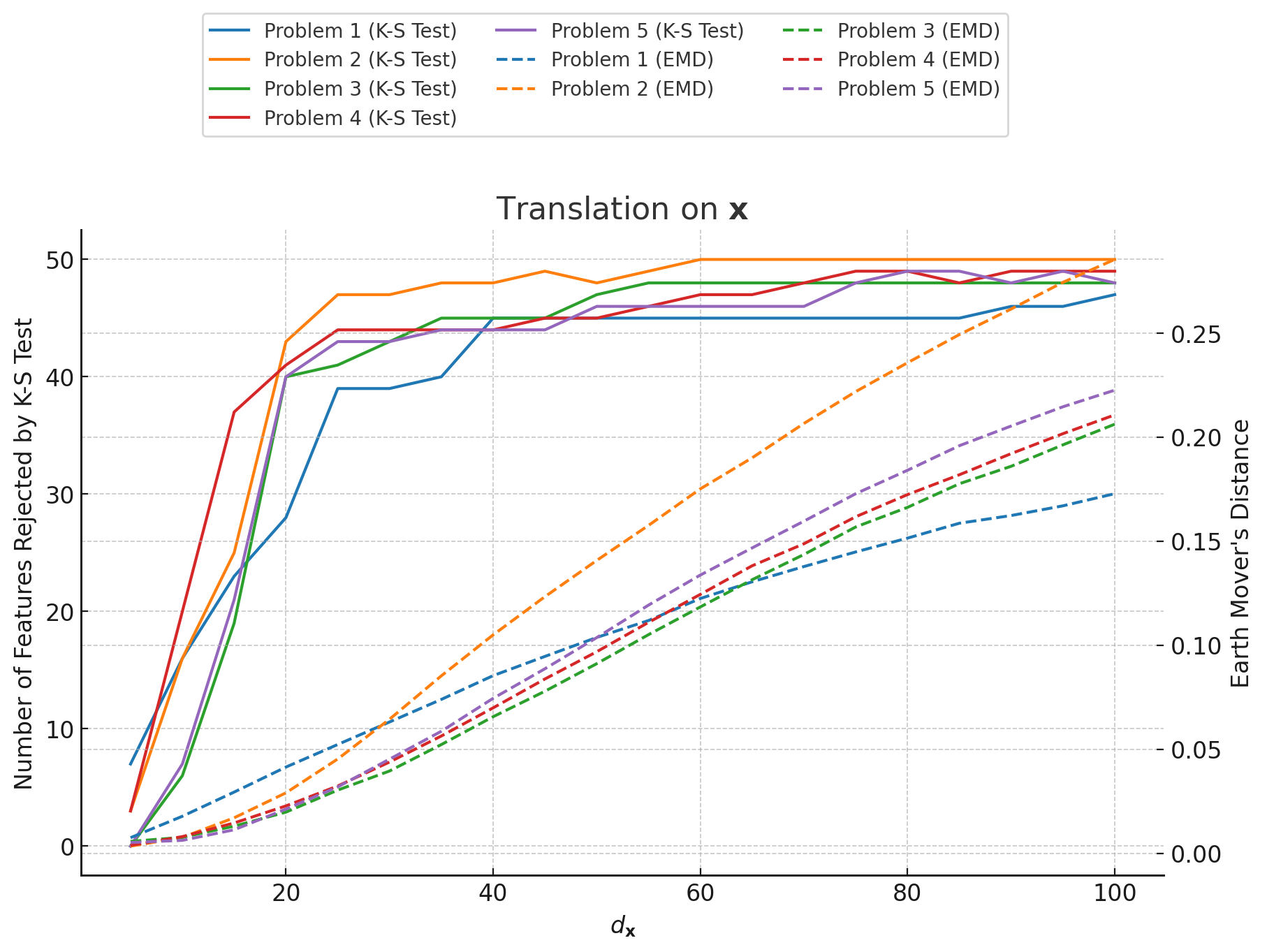}
        \caption{}
        \label{fig:aggregation_x_translation}
    \end{subfigure}
    \begin{subfigure}[b]{0.48\linewidth}
        \includegraphics[width=\linewidth]{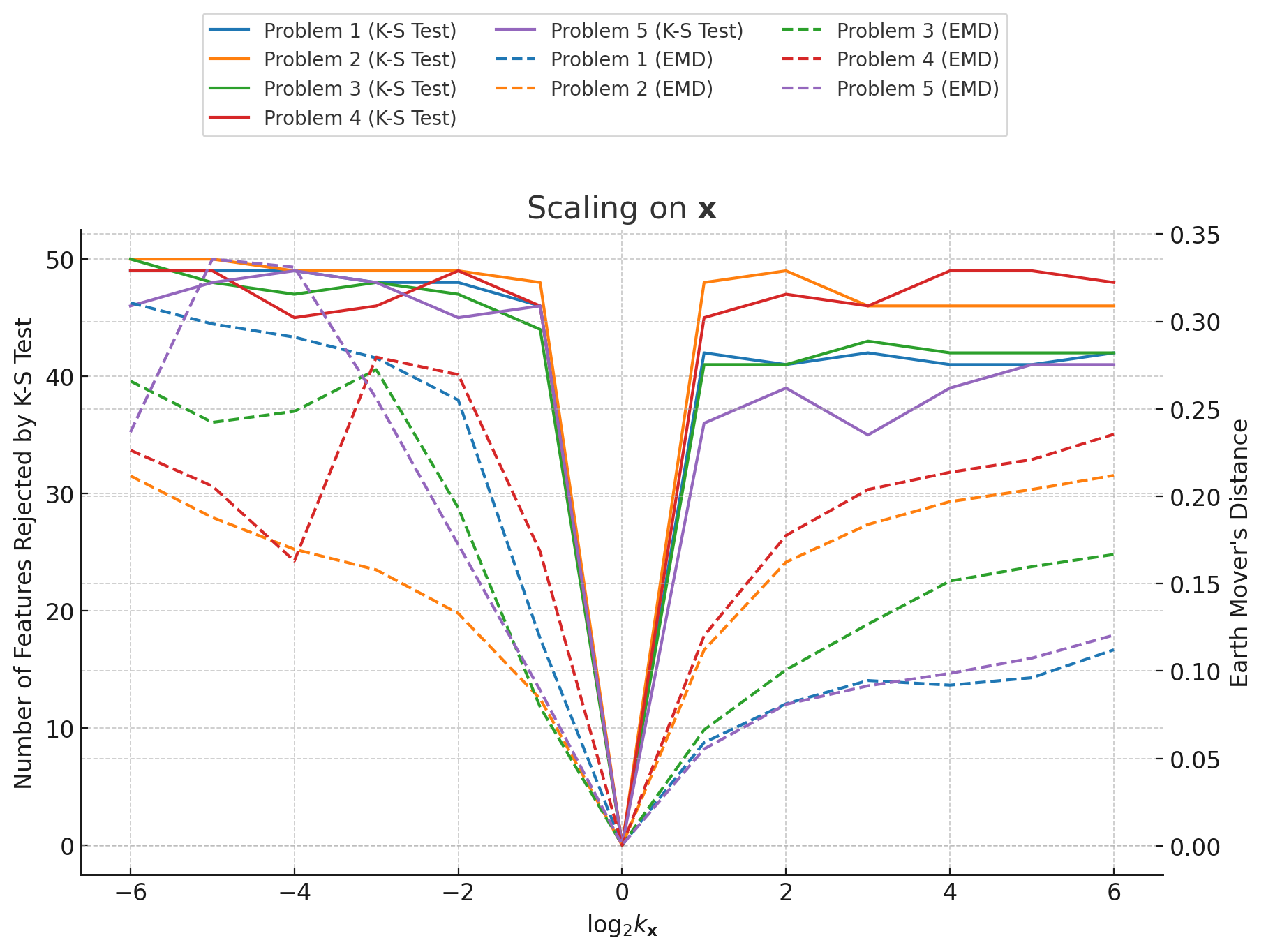}
        \caption{}
        \label{fig:aggregation_x_scaling}
    \end{subfigure}
    \begin{subfigure}[b]{0.48\linewidth}
        \includegraphics[width=\linewidth]{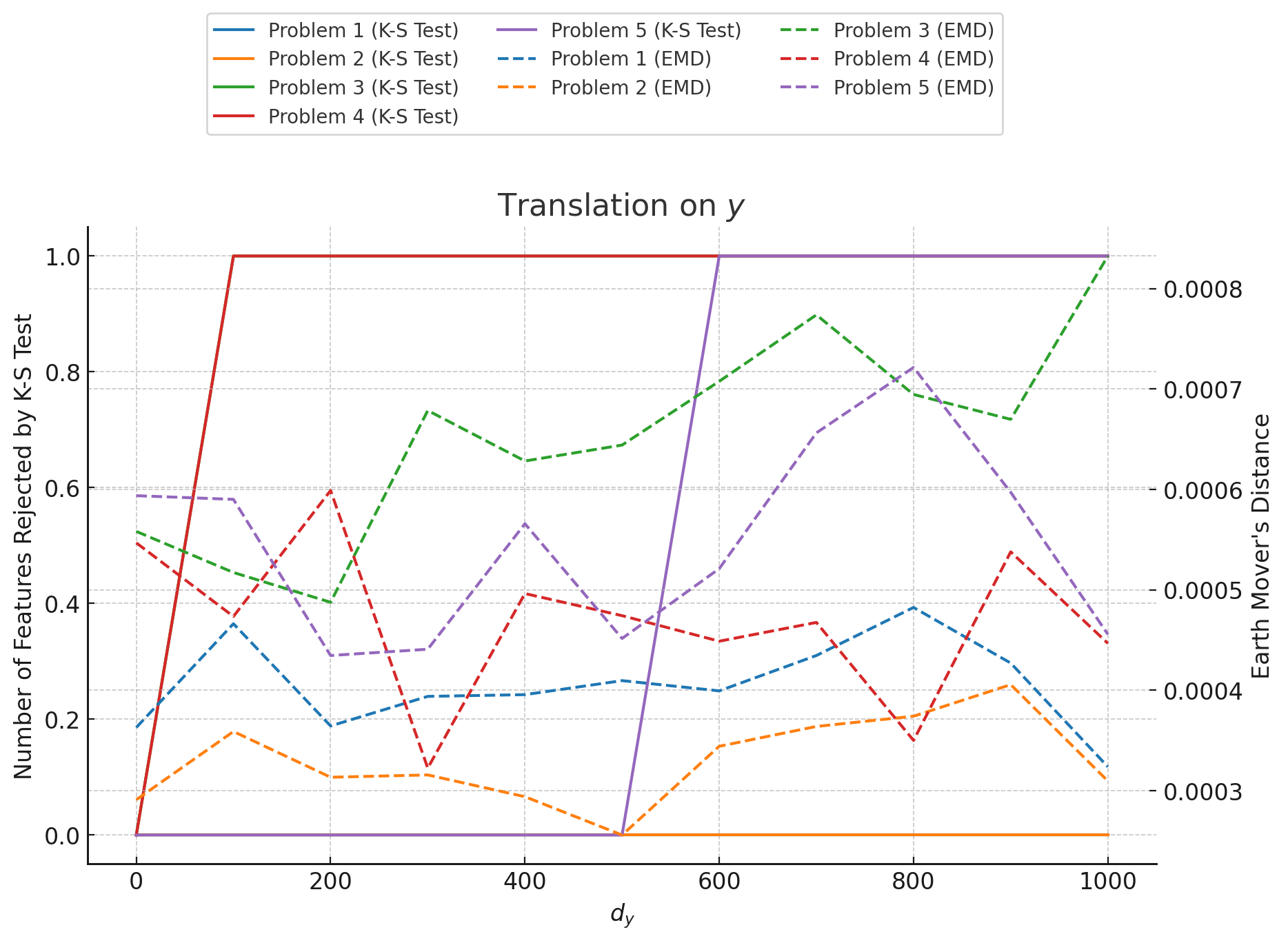}
        \caption{}
        \label{fig:aggregation_y_translation}
    \end{subfigure}
    \begin{subfigure}[b]{0.48\linewidth}
        \includegraphics[width=\linewidth]{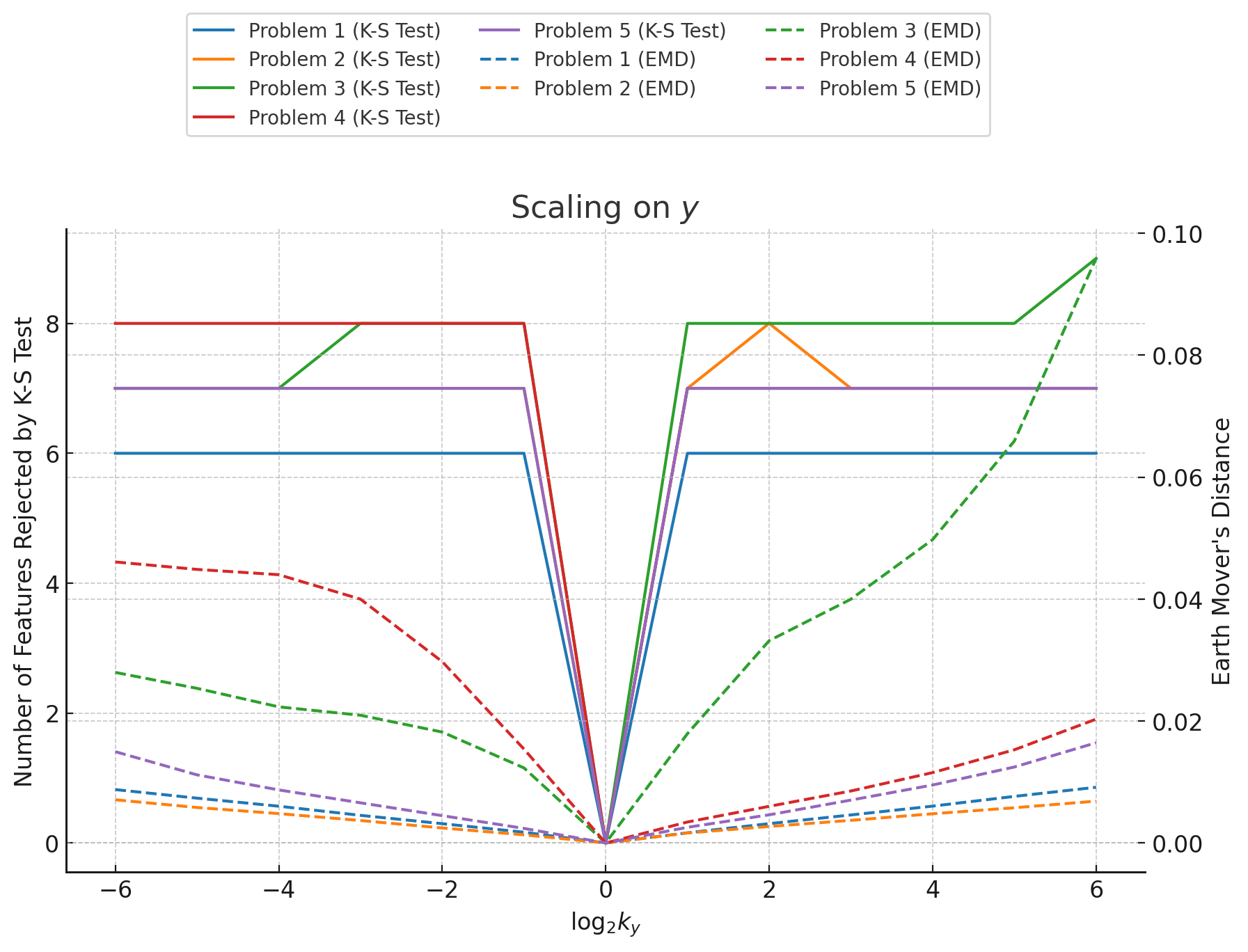}
        \caption{}
        \label{fig:aggregation_y_scaling}
    \end{subfigure}
    \caption{Results of applying four types of transformations to the base functions. EMD (dashed lines) and the number of features (solid lines) rejected by the KS-test between the original and transformed features via translation (left column) or scaling (right column) applied to $\boldsymbol{x}$ (top row) or $y$ (bottom row). Different colours represent different base problems. The EMD results of different problems are calculated based on normalized feature values.}
\end{figure*}

\subsection{Impact of transformations on the search space}
\subsubsection{Translation}
The first transformation method we consider is the translation of the search space. Since we generated translation vectors with varying bounds, we focus on the relation between the chosen bound and the ELA features. As discussed in Section~\ref{sec:methods}, we use both a per-feature KS-test and a distance measure to quantify the changes on the landscape features. The total number of test rejections, as well as the normalized earth mover distance (EMD), is shown in Figure~\ref{fig:aggregation_x_translation}, on the corresponding vertical axes. Note that for each translation limit in this figure, we aggregate over 10 translation vectors independently sampled within the limit. 

Figure~\ref{fig:aggregation_x_translation} shows that the translation factors have a \textit{linear impact} on the overall ELA-feature distribution, as indicated by EMD. For most individual features, the smallest translations don't yet lead to statistically significant changes, but this number quickly increases to almost all features when the translations become larger. In fact, the only features which are \textit{unaffected} by this transformation are those which measure properties of the samples themselves without considering the function values (the PCA-class of features from pflacco). Intuitively, the larger translations can move most of the original function structure outside of the considered box-constrained domain, resulting in a very different function landscape.



\subsubsection{Scaling}
Our scaling-based transformation is parameterized in a similar way to the translation, where we vary the scaling factor logarithmically between $2^{-6}$ and $2^6$. As such, Figure~\ref{fig:aggregation_x_scaling} follows the same structure as the previously discussed Figure~\ref{fig:aggregation_x_translation}, by showing both the change in overall distribution according to EMD and the number of individual features which are statistically significantly impacted by the corresponding rescaling. As opposed to the previous figure, the rescaling is not randomized, and as such, the lines shown represent individual problem instances. A scaling factor of $2^0$ corresponds to the setting of no rescaling, for which we by definition have no change to the base functions. 

In Figure~\ref{fig:aggregation_x_scaling}, we can see that the impact of rescaling is rather \textit{immediate}. Even factors $2^1$ and $2^{-1}$ cause statistically significant changes in almost all ELA features. This is particularly interesting to note on the side of the negative factors, which correspond to zooming in on a smaller part of the function, since this confirms that more local landscape features \textit{vary significantly} from the overall function~\cite{jankovic2019adaptive}, which is an important aspect to consider when basing algorithmic decisions based on ELA features collected over the course of an optimization run.

\subsubsection{Rotation}

For the remaining domain transformations, the ones based on rotation, we don't have a parameterized setup similar to those in the previous subsections. Instead, as described in Section \ref{ExperimentalSetup:SpatialTransformations}, we randomly generated 30 feasible rotation matrices, which means that they are orthogonal matrices distributed in the $[-1, 1]^{10 \times 10}$ space. 
To analyze how rotation influences the ELA features, we introduce $\textit{diff}_{ij}$ as a value to measure the change:
\begin{equation}
    \textit{diff}_{ij} = \left | \frac{\textit{mean}_{0j} - \textit{mean}_{ij}}{\textit{mean}_{0j}} \right | \times 100,
    \label{exp:diff}
\end{equation}

where $\textit{mean}_{ij}$ is the mean value of the $j$-th ELA feature after applying the $i$-th rotation matrix, and $\textit{mean}_{0j}$ is the mean value of the $j$-th ELA feature without rotation.

\begin{figure}
    \centering
    \includegraphics[width=\linewidth]{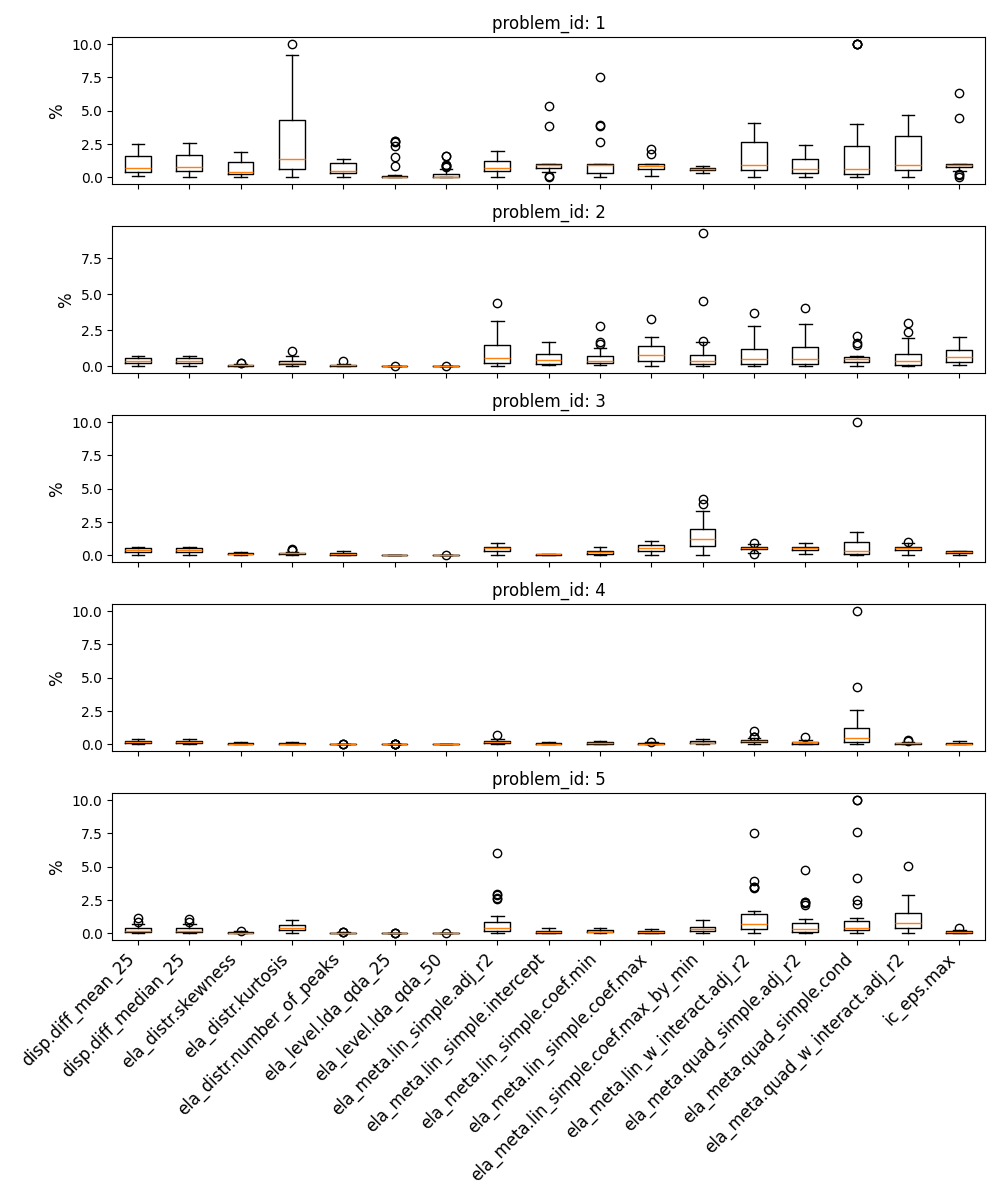}
    \caption{Features of instances rotated on $\boldsymbol{x}$, per problem. Only features where $diff_{ij}\geq 1\%$ on at least one problem are shown. }
    \label{fig:rotate_mean}
\end{figure}

Figure~\ref{fig:rotate_mean} shows the distribution of these relative differences across 30 considered random rotation matrices. To keep the figure readable, we only show ELA-features that have a relative difference of at least $1\%$ on at least one base function. The fact that only 17 features are shown indicates that these random rotations are relatively \textit{less impactful} on the landscape than the search space translation and scaling discussed previously. Of note is that problem 1 shows the \textit{most differences in feature values} between the original and the rotated versions. 

\subsection{Impact of transformations on the objective value}

The impact of transformations on the objective function values has been the point of some discussion since experimental results suggest that not all features are fully invariant to these types of transformations~\cite{vskvorc2022comprehensive}. However, many of the algorithms used within evolutionary computation are comparison-based, and thus not influenced by monotone changes in objective value
. As such, recent studies suggest that function values should be normalized \textit{before applying ELA}, as this would limit the impact of objective value scaling~\cite{prager2023nullifying}. 

To better understand what features are impacted by transformations on the objective value, we again consider parameterized translation and scaling methods. 

\subsubsection{Objective translation}

For translation, we plot the total number of rejections for each translation limit, as well as the overall EMD, in Figure~\ref{fig:aggregation_y_translation}. It should be mentioned that KS rejections for problems 1 and 2 coincide exactly, the same can be said about problems 3 and 4. In this figure, we see that the impact of this transformation is much \textit{smaller} than those on the domain, with only one feature (\textit{ela\_meta.lin simple.intercept}) being statistically significantly different when the applied translation is significantly large for problems 4 and 5. For the remaining problems, the magnitude of the change was not large enough to find statistically significant differences between the translated and original problems, although the continued increase of EMD suggests that with larger transformations this \textit{might} change.



\subsubsection{Objective scaling}

Figure~\ref{fig:aggregation_y_scaling} shows the impact of the scaling transformation on the considered base problems (KS rejections for problems 4 and 5 coincide exactly for positive scaling factors. ). Here, we see a \textit{larger difference} between the original and transformed problems, with up to 8 features being statistically significantly impacted. When comparing the results of this scaling to the transaction, we should note that the scaling factors are much more extreme than the translation limits, which might explain the differences in scale of EMD between Figures~\ref{fig:aggregation_y_translation} and \ref{fig:aggregation_y_scaling}. 


\subsection{Sensitivity of ELA features}
\begin{figure}[!t]
    \centering
    \includegraphics[width=\linewidth]{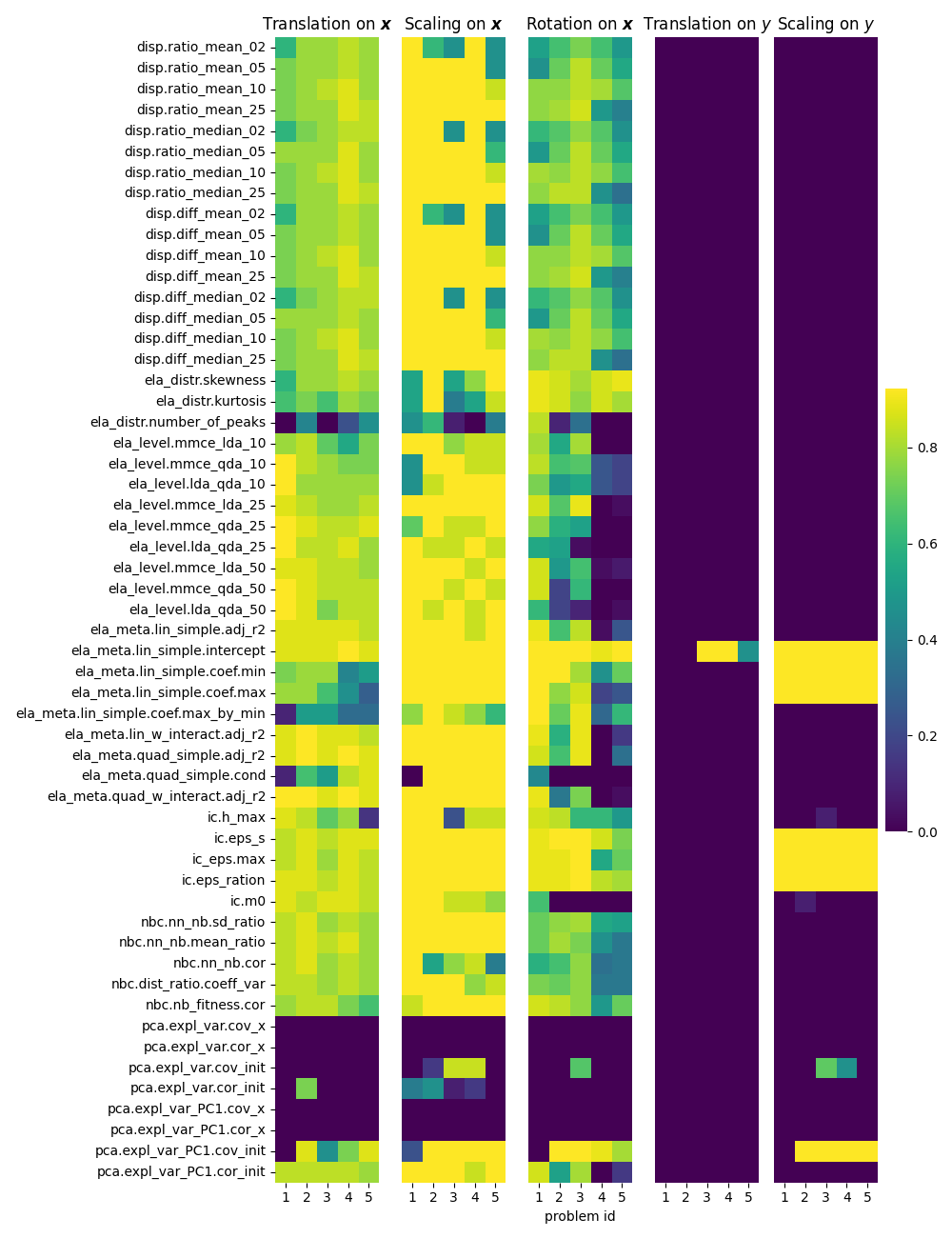}
    \caption{Sensitivity of 55 ELA features after applying 5 types of transformation.
    The brighter the color, the more sensitive the corresponding ELA feature is after this transformation.
    The horizontal axis shows the problem id, while the vertical axis is the ELA feature name. Sensitivity is measured as the fraction of transformed instances in which the distribution of the feature was statistically different according to the KS-test.
    }
    \label{fig:heatmap}
\end{figure}

To get a per-feature view of the impact from the considered transformations, we aggregate the feature changes across all instances created by each transformation method into a measure of \textit{feature sensitivity}. This is calculated as the fraction of transformed instances in which the distribution of the feature was statistically different according to the KS-test. 
        
In Figure~\ref{fig:heatmap}, we show this sensitivity for each feature on each base function. We note that \textit{very few features are fully invariant} to all transformations, with only the PCA-based feature set showing no changes when applying domain transformations. Indeed, the PCA features with no changes at all are those which depend only on the distribution of samples within our domain, which is kept static throughout all instances. On the other hand, the PCA features which include information on the function values do seem somewhat \textit{sensitive}, depending on which underlying function is considered. 

We also observe that, while the intercept of the linear model is the only feature sensitive to our applied function-value translation, with more extreme scaling-based transformations the other coefficient values from the linear model are impacted as well. This is also the case for the \textit{information content features}, which is surprising, given its seemingly robust ability to contribute to algorithm selection models even in the quantum domain~\cite{perez2023analyzing}. 

Finally, we observe some interesting patterns when considering the rotation-based transformations. We see the level-set features on the first 3 functions are much more heavily impacted than on the last 2, possibly suggesting that these functions are impacted significantly by the boundary-effects incurred by rotation - this requires further investigation. 



\section{Conclusion \& Future Work}
\label{sec:conclusion}

In this paper, we have shown that applying transformations to a set of base functions can lead to significant changes in low-level landscape features as measured by ELA. While the impact of transformation methods scales with their disruptiveness, even seemingly small changes to the domain, such as minor translation or simple rotation, have a statistically significant impact on a rather large subset of ELA features. These findings suggest that great care should be taken when designing instance generation mechanisms for the CEC2022 base functions considered here if the aim is to maintain the low-level features present in the current set of functions. 

Another question which remains unanswered is whether we should consider the full set of ELA features going forward. For example, the intercept of a fitted linear model surely contains some information about the landscape, but given that it is highly dependent on the specific range of function values, we can question its use for more general problem feature detection or future algorithm selection. Previous work has suggested that a normalization procedure should be applied to the function values before ELA calculation~\cite{prager2023nullifying}, but this merely shifts the question to, e.g., logarithmic transformations of the function value. 

An overarching question we identify here is how robust the intuitive link is in practice between low-level landscape features, such as ELA, and the intuitive high-level properties which they aim to capture. Many studies using ELA are rather limited in scope, and while they show great performance within benchmarking suites, generalizability to other setups seems rather poor~\cite{vermetten2023ma, kostovska2022per}. More research into the link between high-level landscape properties, ELA features and algorithm behaviour is required to better understand how we can move towards more generalizable results for our automated algorithm selection studies. 



\bibliographystyle{IEEEtran}
\bibliography{citation}

\end{document}